\documentclass{article}

\usepackage[preprint]{neurips_2021}
\usepackage{floatrow}

\usepackage[utf8]{inputenc} 
\usepackage[T1]{fontenc}    
\usepackage{url}            
\usepackage{booktabs}       
\usepackage{amsfonts}       
\usepackage{nicefrac}       
\usepackage{multirow}
\usepackage{caption}
\usepackage{microtype}      
\usepackage{xcolor}         
\usepackage{amsmath}
\usepackage{amssymb}
\usepackage{algorithm}
\usepackage{algorithmic}
\usepackage{amstext}
\usepackage{graphicx}
\usepackage{graphics}
\usepackage{epstopdf}
\usepackage{adjustbox}
\usepackage[colorlinks]{hyperref}
\setcitestyle{round,citesep={,}}

\title{Spectral Propagation Graph Network for Few-shot Time Series Classification}

\author{%
  Ling Yang\textsuperscript{1,2},\ Shenda Hong\textsuperscript{1,2}, Luxia Zhang\textsuperscript{1,2}\\
  \textsuperscript{1}Institute of Medical Technology, Health Science Center of Peking University, Beijing, China\\
  \textsuperscript{2}National Institute of Health Data Science, Peking University, Beijing, China\\
  \texttt{yangling0818@163.com, hongshenda@pku.edu.cn,  zhanglx@bjmu.edu.cn} \\
}
\usepackage{amsmath}

\begin{document}

\maketitle

\begin{abstract}
Few-shot Time Series Classification (few-shot TSC) is a challenging problem in time series analysis. It is more difficult to classify when time series of the same class are not completely consistent in spectral domain or time series of different classes are partly consistent in spectral domain. To address this problem, we propose a novel method named Spectral Propagation Graph Network (SPGN) to explicitly model and propagate the spectrum-wise relations between different time series with graph network. To the best of our knowledge, SPGN is the first to utilize spectral comparisons in different intervals and involve spectral propagation across all time series with graph networks for few-shot TSC. SPGN first uses bandpass filter to expand time series in spectral domain for calculating spectrum-wise relations between time series. Equipped with graph networks, SPGN then integrates spectral relations with label information to make spectral propagation. The further study conveys the bi-directional effect between spectral relations acquisition and spectral propagation. We conduct extensive experiments on few-shot TSC benchmarks. SPGN outperforms state-of-the-art results by a large margin in $4\% \sim 13\%$. Moreover, SPGN surpasses them by around $12\%$ and $9\%$ under cross-domain and cross-way settings respectively.
\end{abstract}

\section{Introduction}
Time Series Classification (TSC) \citep{esling2012time, yang200610} is an important task for many applications across domains like healthcare, financial markets, energy management, traffic system and environmental engineering. In the last decade, deep neural network \citep{fawaz2019deep,kashiparekh2019convtimenet,wang2017time,xu2022cobevt,xu2022v2x} such as Long Short-Term Memory (LSTM) \citep{sundermeyer2012lstm} has been widely used in the areas mentioned above and achieved great success with state-of-the-art results in TSC task. However, when dealing with scarce labeled data (few-shot scenario), the performance of TSC models tend to degrade significantly \citep{fawaz2018transfer,yang2022unsupervised}. Few-shot scenario is inevitable in many of above areas. For example, in healthcare area \citep{rajkomar2018scalable}, it is promising to detect cardiac arrhythmias \citep{huikuri2001sudden} via ECG time series classification \citep{pyakillya2017deep,saritha2008ecg}. But labeled ECG time series are hard to acquire because: 1) some diseases (such as ventricular tachycardia \citep{wellens2000electrical}) rarely happen but are extremely risky; 2) the collection of medical data is inconvenient and costs too much; 3) labeling medical data needs medical expertise and also requires heavy cost.

\begin{figure}
\includegraphics[width=\linewidth]{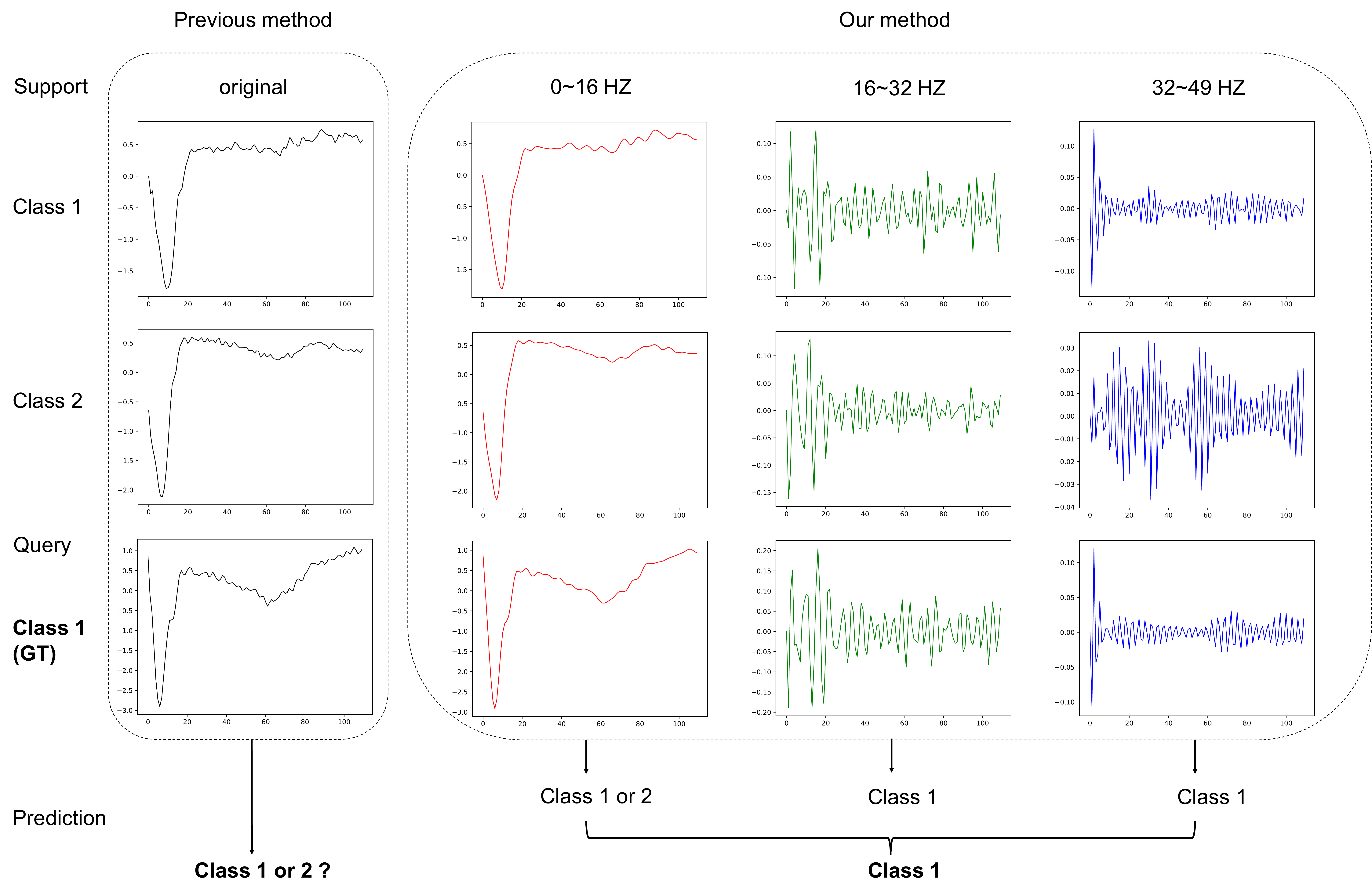}
\caption{Our proposed method utilizes the spectrum-wise comparisons and propagates spectral relations across time series to make final predictions for query time series from a holistic view.}
\label{fig:intro}
\end{figure}

An example of few-shot TSC is shown in Figure \ref{fig:intro}.
Time series in the first two rows are two labeled support samples and time series in the bottom row is query sample which is to be classified.
The ground-truth of query time series is Class 1. We can see that support time series of Class 1 and query time series are dissimilar in $0\sim16$ Hz. In the meanwhile, support time series of Class 2 and query time series are coincidentally similar in $0\sim16$ Hz. It is noted that such differences in low frequency domain might be just caused by baseline wandering distortion and it shouldn't be considered too much in classification. However, such scenario has a significant negative impact on few-shot TSC, which means the training procedure of scarce labeled samples is prone to be seriously affected by such samples and it could bring deviation to the optimization of models. 

However, few recent efforts have been make to promote performance in this few-shot scenario \citep{fawaz2018transfer,kashiparekh2019convtimenet,serra2018towards}. Approaches such as \citep{DBLP:journals/corr/MalhotraTVAS17} focus on transfer learning. They try to solve the problem by pre-training a model on many well-labeled source datasets and fine-tuning it on the target dataset, which may lead to overfitting and be unable to guarantee a satisfied result. Moreover, the performance is hard to guaranteed if the source and target domain have quite different data distributions, which is hard to ensure. For the purpose of model optimization and training efficiency, a specific predictive model for few-shot TSC tasks is desired. MNet \citep{narwariya2020meta} first tries to tackle this problem using meta-learning \citep{vilalta2002perspective} from the perspective of few-shot learning \citep{wang2020generalizing}. It uses meta-learning based pre-training method by the guide of triplet loss function \citep{schroff2015facenet} for few-shot TSC. Nevertheless, its proposed method does not leverage inherent relations within spectrum in a few-shot TSC task. Beyond that, the triplet loss is limited to sample policy selection and outliers may have significant impact on training procedure. In conclusion, these previous methods still fail to deal with the problem of being easily confused by confounders illustrated in Figure \ref{fig:intro}.

In this paper, we aim to maximize the minor differences of various time series and pay more attention to characteristics in common between time series of the same class. We devise SPGN to expand original time series to a range of split time series in power spectral domain for later relations extraction and integrate spectral relations with label propagation for spectral propagation in graph networks. 
The main process of SPGN consists of Spectral Relations Acquisition (SRA) and Spectral Propagation (SP). SRA expands time series with an average split in power spectral space and performs spectrum-wise comparisons to acquire rich spectral relations. SP utilizes spectral relations by integrating them into label propagation and promotes the procedure of SPA in turn. Such an alternate operation of SPA and SP fully exploits the inherent correlated attributes of time series and provides a global view in few-shot TSC tasks, which contributes to significant generalization and transferring abilities in cross-domain few-shot TSC. Furthermore, the architecture of SPGN make it easy for extending it to cross-way few-shot TSC tasks where the number of classes are different between training and evaluation.
Our main contributions are summarized as follows:
\begin{itemize}
\item To the best of our knowledge, SPGN is the first to \textbf{explicitly model spectrum-wise relations in few-shot TSC}, which expands the minor differences of diverse time series and highlights the major similarities of same time series in spectral domain. The further ablation study illustrates the effectiveness of spectral comparisons in different intervals. 
\item SPGN firstly attempts to \textbf{adopt the graph network architecture for better combination of spectral relations acquisition and label propagation in few-shot TSC}, which helps in promoting spectral propagation and significantly improves the final prediction.
\item In extensive experiments on popular benchmark datasets for few-shot TSC tasks, SPGN achieves a state-of-the-art result and has a significant improvement of \textbf{4\%$ \sim $13\%} on average accuracies over previous work. Additionally, our algorithm outperforms existing few-shot TSC methods by \textbf{12\% and 9 \%} under cross-domain and cross-way settings respectively, which proves the superior generalization and transferring abilities of SPGN.
\end{itemize}

\section{Related Work}
\label{rel_work}
\paragraph{Time series classification}
With the significant increase volume of temporal data, a large number of TSC algorithms have been proposed in recent years \citep{esling2012time,fawaz2019deep,yang200610,yang2022unsupervised-arxiv}. To name a few, Dynamic Time Wrapping (DTW) \citep{bagnall2017great} provides a strong baseline with the dynamic time warping distance. A popular traditional TSC method incorporates a Nearest Neighbor (NN) classifier coupled with a distance function \citep{lines2015time}.  BOSS \citep{schafer2015boss} forms a discriminative bag of words to promote the performance in TSC tasks. Due to the success of Deep Learning \citep{lecun2015deep} in various classification tasks, several discriminative deep learning methods have been proposed to solve the TSC task. Hand engineering approaches \citep{hatami2018classification,wang2015imaging,wang2015spatially} transform time series data into image-like input for feature extraction, which is mostly inspired by computer vision research. As opposed to feature engineering, end-to-end models aims to incorporate the feature learning process while fine-tuning the discriminative classifier \citep{nweke2018deep}.

\paragraph{Few-shot learning}
Few-shot learning is to predict unlabeled samples (query set) given few labeled samples(support set), which has important worth of research and application. Finetuning is a traditional method to train a predictive model for few-shot learning in practice \citep{chen2018closer, nakamura2019revisiting}.
The method of meta-learning \citep{finn2017model} introduces the concept of episode to solve the few-shot classification task explicitly and designs a gradient-based optimizer for fast adaptation of model parameters when encountering novel samples.
An episode is one iteration of model training, where a batch of few-shot tasks are sampled for optimization. 
Under the framework of meta-learning, metric learning methods \citep{snell2017prototypical,sung2018learning,vinyals2016matching} focus on optimizing feature embedding of input data in few-shot tasks. 
\paragraph{Graph network}
Graph network was designed for processing graph-structured data at first \citep{gori2005new,scarselli2008graph} and it mainly refines the node feature by aggregating neighboring nodes and update itself recursively \citep{pmlr-v162-yang22d}.
Transductive Inference \citep{joachims1999transductive} was first introduced to minimize errors of a particular test dataset and it shows a massive advantage over inductive inference. Graph-based methods \citep{kim2019edge,liu2018learning,satorras2018few,yang2020dpgn} are suitable for involving transduction to conduct label propagation, which transfers labels from labeled samples to unlabeled samples effectively. All these prior works show great efficiency of the combination of graph network and transductive inference \citep{wang2007label,zhou2004learning,yang2022omni} in classification tasks. 
\section{The Proposed Approach}
\label{approach}

In this section, we first introduce the problem definition of few-shot TSC tasks.
And then we provide detailed explanation of our proposed algorithm.

\subsection{Problem Definition}
\label{problem definition}
The goal of few-shot TSC tasks is to learn a predictive model that generalizes well when classifying time series in the case where only few labeled time series are provided.
Specifically, each few-shot TSC task contains a \emph{support set} $\mathcal{S}$ and a \emph{query set} $\mathcal{Q}$, which are composed of time series. In the training stage provided training data $\mathbb{D}^{train}$, 
the support set $\mathcal{S} \subset \mathbb{D}^{train}$ contains $N$ classes with $K$ time series for each class (i.e., the $N$-way $K$-shot setting), which can be denoted as $\mathcal{S} = \{(x_{_1},y_{_1}),(x_{_2},y_{_2}),\dots,(x_{{_N} {_\times} {_K}},y_{{_N} {_\times} {_K}})\}$.
The \emph{query} set $\mathcal{Q} \subset \mathbb{D}^{train}$ has $M$ samples and can be denoted as $\mathcal{Q} = \{(x_{_{N \times K+1}},y_{_{N \times K+1}}),\dots,(x_{_{N \times K+{M}}},y_{_{N \times K+{M}}})\}$.
Labels of time series are provided for both support set $S$ and query set $Q$.
In the evaluation stage provided testing data $\mathbb{D}^{test}$, our goal is to train a predictive model that could map the query time series $\mathcal{Q} \subset \mathbb{D}^{test}$  to the corresponding label accurately with few support samples from $\mathcal{S} \subset \mathbb{D}^{test}$. Labels of support time series and query time series are mutually exclusive.
Similar to few-shot image classifications \citep{finn2017model}, few-shot TSC adopts the episodic training schema to learn model parameters, which is consistent with evaluation stage. An iteration of training procedure is denoted as an episode and contains a batch of few-shot TSC tasks sampled from $\mathbb{D}^{train}$. Such a episodic training method is designed for promoting generalization ability of few-shot learning models. 


\begin{figure}[h]
\centering
    \includegraphics[width=14cm]{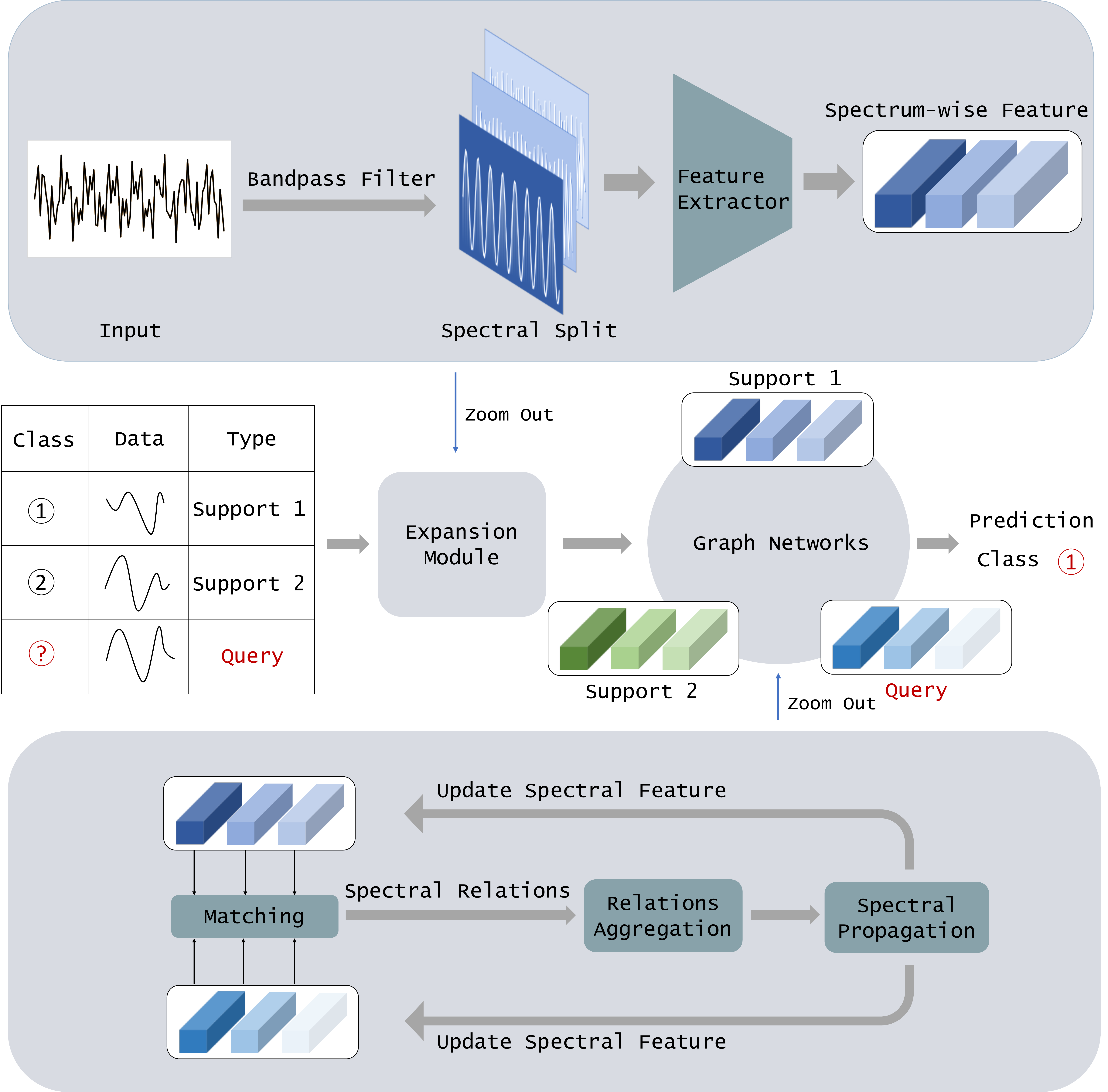}
    \vspace{1em}

\caption{The overall framework of SPGN. A 2way-1shot few-shot TSC task is illustrated. First, Support and query time series are all delivered to expansion module for acquiring spectrum-wise feature. Then the obtained features are used to construct graph network, where each node represents a time series instance. Finally, SPGN alternately runs Spectral Relations Acquisition and Spectral Propagation to make prediction for query time series.}
\label{fig:gnn}
\end{figure}

\subsection{Spectral Propagation Graph Network}
\label{spgn}

In this subsection, we will introduce the proposed spectral propagation graph network in detail. As illustrated in Figure \ref{fig:gnn}, SPGN contains a feature extractor and a graph network. The feature extractor is a base component for extracting temporal patterns like previous work \citep{kashiparekh2019convtimenet}. The graph network architecture $G = (V, E)$ we devise plays an important role in our algorithm, which performs two main processes, namely Spectral Relations Acquisition and Spectral Propagation. SPGN conducts the processes of SRA and SP alternately to update nodes $V^l$ and edges $E^l$ in graph $G^l$, where $l$ represents the $l$-th graph network. SPGN outputs the final prediction for query time series after the processing of several graph networks. Furthermore, the processes of SRA and SP in $G^l$ both utilize the previous feature in $G^{l-1}$  for update, which enhance the stability of training procedure and ensure that the information flow between label and spectrum is still efficient across several graph networks.
\subsubsection{Spectral Relations Acquisition}
\paragraph{Expansion of time series in spectrum}
\label{split}
Previous methods \citep{kashiparekh2019convtimenet,narwariya2020meta} usually capture temporal patterns of the whole time series when encountering few-shot TSC tasks. They may easily over-highlight the minor differences of same class and be confused by shared feature of different classes. To reduce the impact of confounders, SPA conducts an average split of time series in power spectral space for fine-grained feature and makes spectrum-wise comparisons for global information when judging whether two time series are of same class. Therefore, each split time series contain the same amount of spectral information. Thus, a bandpass filter based on the Fast Fourier Transform (FFT) and inverted FFT is employed on each time series to get a group of spectrum-wise time series. We first computes the power spectral density (PSD) of the time series. The PSD set with corresponding spectrum set can be defined as follows:
\begin{align}
    PSD = \frac{abs(FFT(x))^2}{N}
\end{align}
where $N$ denotes the number of sampling points and $x$ is the input time series. $FFT$ is the Fast Fourier Transform. Then 
we accumulating PSD for each frequency scale as follows:
\begin{align}
    H(f_i) = \sum_{i=1}^{n}(PSD(f_i) \times f_i)
    \label{accumulate}
\end{align}
where $n \in \mathbb{Z}$ and $0\leq n<\frac{f_w}{2}$.
Given total power spectral $I_{total}=\sum_{i=1}^{\frac{f_w}{2}}(PSD(f_i) \times f_i)$ of time series, we perform average split on spectrum to get $s$ time series. The process is specified as follows:
\begin{align}
    f_j = H^{-1}((\frac{I_{total}}{s})*j)
\end{align}
where $H^{-1}$ is the reverted function of equation \ref{accumulate}. $f_j$ is the $j$-th split frequency scale and the spectral interval of $j$-th split time series is $[f_j, f_{j+1}]$. Then the bandpass filter $B$ is applied for acquire a set of spectrum-wise time series with the above spectral intervals. The $j$-th spectrum-wise time series $x_j$ is computed as:
\begin{align}
    x_j = B(x, f_j, f_{j+1})
\end{align}

where $x$ is the original time series, $f_j$ and $f_{j+1}$ are lower and upper bounds of the bandpass filter respectively. 

Each group of processed support and query time series is first delivered to the feature extractor for extraction of basic temporal patterns. For fair comparison, SPGN use the ResNet-like architecture \citep{he2016deep} for feature extraction as illustrated in \citep{hong2020holmes}. We slightly adapt the original ResNet to the model suitable for 1-dimensional data, which is called ResNet-1D. ResNet-1D is composed of four residual blocks and each block contains several residual layers.The last layer of ResNet-1D is followed by a fully-connected layer to project the intermediate feature to the final 128-dimensional feature embedding for each split time series.  

\paragraph{Spectrum-wise comparisons} To make prediction for query time series, SPGN involves graph network to build a fully-connected architecture for element-wise matching. Each node in the initial graph $G^0 = (V^0, E^0)$ is initialized with several feature embeddings from different spectral ranges of the same time series. We do not display the processing of the feature obtained from original unprocessed time series in the latter equations for simplicity and the processing is exactly same with that of spectrum-wise feature.  For $s$-th spectrum, the spectral relation between nodes is defined as follows: 
\begin{align}
    r^0_{ij,s} = {F_{G^0_{rel}} (({v^0_{i,s}-{v^0_{j,s}})^2)}}\,,
    \label{eq1}
\end{align}
where $v^0_{i,s} \in \mathbb{R}^{128}$ and $v^0_{j,s} \in \mathbb{R}^{128}$ are the feature embedding in $s$-th spectrum of $i$-th and $j$-th time series respectively, $F_{G^0_{rel}}$ is the transform network that calculates the spectrum-wise similarity. $F_{G^0_{rel}}$ contains two Conv-BN-ReLU \citep{agarap2018deep,ioffe2015batch,krizhevsky2012imagenet} blocks with the parameter set ${\theta}_{G^0_{rel}}$ and a sigmoid layer.

After capturing $s$ relations in spectral domain between each node pair, SPA concatenates and squeezes all the relations to compute a global relation to represent the edge in $G_0$. Given relation set between $i$-th and $j$-th time series ${R^0_{ij}} = \{r^0_{ij, 0}, r^0_{ij, 1}, \cdot, r^0_{ij, s}\}$, the edge $e^0_{ij} \in \mathbb{R}$ is computed as follows:
\begin{align}
    e^0_{ij} =  \delta(y_i, y_j) \cdot {F_{G^0_{e}} (concat(r^0_{ij, 0}, r^0_{ij, 1}, \cdot \cdot \cdot, r^0_{ij, s}))}\,,
    \label{eq:2}
\end{align}
where $F_{G^0_{e}}$ is composed of several FC-ReLU blocks with the parameter set ${\theta}_{G^0_{e}}$ and a sigmoid layer. $\delta$() is the Kronecker  delta function which outputs $1$ if $y_{i} =y_{j}$ otherwise $0$ ($y_{i}$ and $y_{j}$ are labels of $i$-th and $j$-th time series respectively) and the value would be set to $0.5$ if query time series is one of $i$-th and $j$-th time series. With this process, SPGN completely integrates the spectral relations with label information for latter spectral propagation. It makes the model to adaptively learn which spectrum to concentrate on when comparing time series and correspondingly update spectral feature for better comparisons.
And the edge $e^l_{ij}$ in $l$-th graph network $G^l$ be updated as follows:
\begin{align}
    e^l_{ij} = e^{l-1}_{ij} \cdot {F_{G^l_{e}} (concat(r^l_{ij, 0}, r^l_{ij, 1}, \cdot \cdot \cdot, r^l_{ij, s}))}\,,
    \label{eq:3}
\end{align}
where $F_{G^l_{e}}$ has a same architecture with $F_{G^0_{e}}$. In order to convey edge information from a holistic view in the graph $G^l$, a normalization operation is conducted on the ${e^l_{ij}}$.  
\subsubsection{Spectral Propagation}
Spectral Propagation is devised to involve label propagation with spectral relations through transduction. The edge produced from SRA encoding the rich spectral relations of rime series pairs and it is consistent with relative relationship of labels, which means two series of strong spectral correlation should probably be of the same class. SP exploits the bi-directional effects between spectral relation calculation and label propagation. More accurate spectral relations provide more abundant global information for label propagation. More efficient label propagation contributes to the better discrimination of spectrum-wise feature, which is beneficial for computing spectral relations in turn. Given the feature set $\{v^{l-1}_{0, s}, v^{l-1}_{1, s}, \cdot \cdot \cdot, v^{l-1}_{N \cdot K+M, s}\}$ in $G^{l-1}$ and the edges  $ E^l = \{e^{l-1}_{0}, e^{l-1}_{1}, \cdot \cdot \cdot, e^{l-1}_{N \cdot K+M}\}$, the $i$-th node feature of the $s$-th spectrum can updated as follows: 
\begin{align}
     v^l_{i,s} =  F_{G^l_{v}} (concat(\sum_{j=1}^{N \cdot K+M} (e^{l-1}_{ij}\cdot v^{l-1}_{j, s}), v^{l-1}_{i, s}))\,,
     \label{eq:4}
\end{align}
$F_{G^l_{v}}$ consists of two Conv-BN-ReLU blocks with the parameter set ${\theta}_{G^l_{v}}$. Through this process of label propagation and feature aggregation, the global information produced by SRA are encoded in updated node feature of each spectrum for all time series.

The prediction of each time series can be computed by feeding the corresponding edges in the last graph network $G^{last}$ in SPGN into softmax function:
\begin{align}
    P( \hat{y_i} | x_i)  = {Softmax}  (\sum_{j=1}^{N\cdot K} {e^{last}_{ij}} \cdot \eth (y_j))\,,
\end{align}
where $P( \hat{y_i} | x_i)$ is prediction's probability distribution over all support classes given sample $x_i$, and $y_j$ is the label of $j$th sample in the support time series. $\eth(\cdot)$ represents the one-hot mapping for labels and $e^{last}_{ij}$ stands for the edge. $N \cdot K$ stands for the number of support time series in a few-shot TSC task. To fully exploit spectral relations when optimizing SPGN, we add constraints on every graph networks. The total loss fuction is defined as follows:
\begin{align}
    \mathcal{L}_{SPGN} = \sum_{i=1}^{T} \mathcal{L}_{CE}(P( \hat{y_i} | x_i) , {y_i})\,,
\end{align}
where $T$ stands for the number of graph networks existing in SPGN. $\mathcal{L}_{CE}$ is the cross-entropy loss function and $P(\cdot|\cdot)$ is the prediction probability distribution over all support classes of split query time series $x_i$. ${y_i}$ is the ground-truth label. 
\section{Experiments}
\label{exp}

\subsection{Setups and Main Results}
 \paragraph{Datasets and Protocols}
 \label{dataset}
 We use the public UCR Time Series Classification Archive dataset \citep{dau2019ucr} which is widely used in time series classification to construct few-shot TSC tasks and run evaluation on it. The UCR dataset contains different kinds of TSC datasets with each dataset belonging to a TSC task. For fair comparisons, we follow \citep{narwariya2020meta} and 41 datasets of UCR Archive are used for few-shot TSC tasks. We conduct training and evaluation processes on every dataset. Given a dataset, we randomly sample many few-shot TSC tasks from original train/test splits in UCR Archive. In the training stage, we use Adam optimizer \citep{kingma2015adam} on a single Nvidia RTX3090 GPU with CUDA 11.0 in all experiments with the initial learning rate of 1e-3 and decay the learning rate by 0.5 per 30 epochs. The weight decay is set to 1e-5. After training procedure, we report the mean accuracy (in \%) of evaluation as well as the 95\% confidence interval, which is consistent with previous approaches \citep{narwariya2020meta}. 

\paragraph{Main results} We compare the performance of the proposed model SPGN with several state-of-the-art models, and SPGN surpasses them by a significant margin on all benchmark datasets. Considering that datasets of the UCR Archive consists of the different number of classes which can vary from 2 to 50, we only fix the number of labeled samples in each class and conduct experiments based on a $N$-way $5$-shot learning. Following the datasets setting of previous method \citep{narwariya2020meta}, the number of classes $N$ in a $N$-way $K$-shot few-shot TSC task vary across different datasets. As shown in Table \ref{5shot}, the evaluation accuracy of SPGN gets an improvement of $4\% \sim 13\%$ on all datasets for $5$-shot TSC tasks when it is compared to previous work. The promising results reveal the generalization ability of our proposed model and prove the necessity of utilizing spectral relations comparison in few-shot TSC tasks.

\begin{table}
  \caption{Comparisons of $N$-way $5$-shot TSC accuracies on 41 UCR datasets. $N$ varies according to the number of classes across datasets.}
  \label{5shot}
  \centering
  \setlength{\tabcolsep}{0.50mm}
  \begin{tabular}{lllllllllll}
    \toprule
    Dataset     &  DTW & BOSS & MNet & \textbf{SPGN} &Dataset     &  DTW & BOSS & MNet & \textbf{SPGN}  \\
    \midrule
    50words   &64.4 &49.9& 59.1 &\textbf{72.3$\pm$0.54}& InsectW.B.Sound& 47.3 &39.8&48.7&\textbf{56.0$\pm$0.32}\\

    Adiac & 54.0& 70.9& 67.4& \textbf{77.6$\pm$0.41}& Meat& 91.9& 87.6&89.0&\textbf{99.1$\pm$0.17} \\
    Beef  &62.6& 70.1 &65.3&\textbf{83.5$\pm$0.45}& MedicalImages  & 67.5 &48.8&59.2&\textbf{78.2$\pm$0.39}\\
    BeetleFly& 61.4& 78.9  &95.8&\textbf{98.9$\pm$0.18} &Mid.Phal.O.A.G&55.8&47.8&54.7&\textbf{64.9$\pm$0.21} \\
    BirdChicken & 49.6& 92.1  &\textbf{100.0}&\textbf{99.7$\pm$0.17}& Mid.Phal.O.C&55.0&52.6&53.1&\textbf{66.4$\pm$0.34} \\
    Chlor.Conc. & 33.8& 35.6 & 33.1&\textbf{43.8$\pm$0.15} & Mid.Phal.TW &33.9&34.8&35.3&\textbf{45.6$\pm$0.20}\\

    Coffee& 91.4 &97.7 & 97.8& \textbf{99.6$\pm$0.18}& PhalangesO.C&53.5&51.2&53.9&\textbf{61.4$\pm$0.41}\\
    Cricket\_X & 56.7& 49.1&59.4&\textbf{67.3$\pm$0.25}& Prox.Phal.O.A.G& 71.9&73.1 &69.7 &\textbf{80.9$\pm$0.57} \\
    Cricket\_Y& 55.6& 46.1 & 56.2&\textbf{63.7$\pm$0.36}& Prox.Phal.O.C &62.6 &64.5 & 63.8 &\textbf{74.5$\pm$0.41}\\
    Cricket\_Z & 56.0 &48.1& 59.8&\textbf{69.1$\pm$0.28}& Prox.Phal.TW&44.5&41.9&43.2&\textbf{52.0$\pm$0.14}\\
    Dist.Phal.O.A.G& 69.8 &65.8 &70.5&\textbf{78.3$\pm$0.44}& Strawberry&67.1&71.4&75.5&\textbf{82.2$\pm$0.59}\\
    Dist.Phal.O.C&58.3 & 57.5& 58.8&\textbf{65.9$\pm$0.30} & SwedishLeaf&69.0&77.6&77.8&\textbf{88.1$\pm$0.46}\\
    Dist.Phal.TW&44.8 &43.7 &48.1 &\textbf{56.4$\pm$0.21} & synthetic\_control&95.8&86.7&97.1&\textbf{99.7$\pm$0.22}\\
    ECG200& 75.5&72.8 &75.8&\textbf{80.7$\pm$0.33} &Two\_Patterns&97.0&69.2&83.1&\textbf{99.6$\pm$0.21}\\
    ECG5000&49.4&53.3 &54.8 &\textbf{61.4$\pm$0.19} & uWave\_X&61.5&47.9&60.6&\textbf{69.0$\pm$0.32}\\
    ECGFiveDays &66.6 &90.9 &93.9&\textbf{98.7$\pm$0.27} &uWave\_Y&51.8&36.3&47.8&\textbf{60.1$\pm$0.34}\\
    ElectricDevices&42.2 &35.1 &38.0 &\textbf{51.3$\pm$0.29} &uWave\_Z&55.1&48.9&59.9&\textbf{66.7$\pm$0.42}\\
    FaceAll&76.4 &79.5 &78.5 &\textbf{85.2$\pm$0.31} &wafer&92.2&93.6&89.4&\textbf{97.4$\pm$0.35}\\
    FaceFour&86.9&\textbf{100.0} &95.8 &\textbf{99.8$\pm$0.13} & Wine&49.3&57.1&63.1&\textbf{72.3$\pm$0.43}\\
    FordA&54.1 &69.3 &79.7 &\textbf{85.8$\pm$0.40} &yoga&52.5&54.8&54.6&\textbf{63.5$\pm$0.37}\\
    FordB&53.5&58.5&78.7&\textbf{87.1$\pm$0.41} \\

    \bottomrule
  \end{tabular}
\end{table}
\subsection{Generalization and Transferability}
To assess the performance of proposed model from different aspects, we design cross-domain and cross-way experiments comparing SPGN with other methods to validate the generalization and transferability of models. Besides, generalization and transferring abilities of models are vital for few-shot TSC tasks because it is quite common that the settings in training and testing stage are diverse in practical scenarios. 
 \paragraph{Cross-domain evaluation}
 Previous methods fail to adapt to domain shift where $A\xrightarrow{}B$ implies that the model is trained on dataset $A$ but evaluated on dataset $B$. We conduct the cross-domain evaluation, where source and target domains are different, and results are shown in Table \ref{cross}. SPGN is able to quickly adapt to a novel class and is less affected by domain shift between the source and target domains, which is obviously better than other methods. It is noted that the exploitation of spectral information becomes more important as the domain difference grows larger.
 
 \paragraph{Cross-way evaluation} There usually exsists the senario where the number of few-shot classes for evaluation does not match to the one used for training. Previous methods are less conductive because of its architecture. In contrast, our SPGN is more flexible and effective to be extended to modified few-shot setting without re-training of the model. Therefore, we slightly modify other models to be able to make comparisons with SPGN in cross-way evaluation. As shown in Table \ref{cross}, SPGN outperforms the previous methods by a large margin in cross-way few-shot TSC tasks, which has proved its good capabilities of generalization and transferability.
 
\begin{table}
  \caption{Cross-domain and cross-way results. Cross-domain evaluation is conducted on $source \xrightarrow{} target$ dataset pairs and Cross-way evaluation is experimented on FaceAll dataset.}
  \label{cross}
  \centering
  \setlength{\tabcolsep}{0.65mm}{
  \begin{tabular}{cccccccc}
  \toprule
\multirow{2}{*}{Methods}  & \multicolumn{3}{c}{\textbf{Cross-domain}} & \multicolumn{1}{c}{\space}& \multicolumn{2}{c}{\textbf{Cross-way}}\\ \cline{2-4} \cline{6-7} &Adiac $\xrightarrow{}$ ECG200 & FaceAll $\xrightarrow{}$ FordB &50words $\xrightarrow{}$ Beef &&2way $\xrightarrow{}$ 5way& 5way $\xrightarrow{}$ 10way \\ \midrule
DTW & 60.3&42.5&50.2&&62.5&65.3\\
BOSS & 58.1&46.7&59.3&&64.8&69.9\\
MNet & 62.9&69.4&56.8&&65.6&67.1\\
SPGN &\textbf{73.8}&\textbf{81.3}&\textbf{75.0}&&\textbf{73.0}&\textbf{78.4}\\
\bottomrule
 \end{tabular}}
 \end{table}
\section{Analysis}
\label{analysis}
\subsection{The Effectiveness of SPGN}
Our proposed SPGN is able to address the problem that when time series of the same class are not completely consistent in spectral domain or time series of different classes are partly consistent in spectral domain. To support our claim,
we make a statistical analysis about the number of different spectrums in the wrong classification pairs by using SPGN and previous methods as in Table \ref{diff spectrum}. We can find that most of the wrong classification pairs have more than two different spectrums, which percentage largely surpasses other methods. This phenomenon demonstrates that our SPGN is less affected by partial indistinguishable spectrums. Besides, we use SPGN to classify the wrong pairs produced by other methods. From Table \ref{corrected}, we find that our SPGN corrects a large proportion of wrongly classified pairs with 1 or 2 different spectrums, which further prove the effectiveness of SPGN.

\begin{table}
    \centering
        \caption{The distribution percentage of different spectrums in wrongly classified pairs.}
    \label{diff spectrum}
      \setlength{\tabcolsep}{1.mm}{
    \begin{tabular}{c|cccc}
    \toprule
Count of Spectrums&	DTW&	BOSS	&MNet&	\textbf{SPGN}\\
\midrule
1&53\%&62\%&58\%&\textbf{11\%}\\
2&34\%&28\%&31\% &\textbf{15\%}\\
$\geq$ 2&13\%& 10\%&	11\%&\textbf{74\%}\\
\bottomrule
\end{tabular}
}
\end{table}
\begin{table}
    \centering
        \caption{The corrected proportion of wrongly classified pairs produced by other methods with using SPGN.}
    \label{corrected}
      \setlength{\tabcolsep}{1.mm}{
    \begin{tabular}{c|ccc}
    \toprule
Corrected Proportion &	DTW&	BOSS	&MNet\\
\midrule
1&\textbf{91\%}&\textbf{85\%}&\textbf{93\%}\\
2&\textbf{83\%}&\textbf{88\%}&\textbf{79\%}\\
$\geq$ 2&67\%&42\%&57\%\\
\bottomrule
\end{tabular}
}
\end{table}
\paragraph{Robustness to Noise}
Our SPGN is robust to the noise since we split the time series in spectral domain and it reduces the influences of noise with the spectral propagation. We conduct experiments with high, middle and low frequency noise and results are listed in Table \ref{noise}. When the noise is added, the accuracy reduction of SPGN is below 1\% while other models have a large accuracy reduction about 3\%. Therefore our SPGN is more robust to noises of different frequencies comprared with other methods.
\begin{table}[ht]
    \centering
        \caption{Accuracy comparisons with different noises.}
    \label{noise}
    \setlength{\tabcolsep}{1.0mm}{
    \begin{tabular}{l|lll}
    \toprule
High Frequency Noise &	DTW	&MNet&SPGN\\\midrule
ECG5000 (w/o)&	49.4&	54.8&61.4\\
ECG5000 (w/)&	45.8 (-3.6)&	51.2 (-3.6)&	60.5 (\textbf{-0.9})\\
Cricket\_Z (w/o)&	56.0&	59.8&69.1\\
Cricket\_Z (w/)&	52.4 (-3.6)&	56.0 (-3.8)&	68.3 (\textbf{-0.8})\\
Prox.Phal.TW (w/o)&	44.5&	43.2&52.0\\
Prox.Phal.TW (w/)&	40.7 (-3.8)&	40.8 (-2.4)&	51.1 (\textbf{-0.9})\\
\bottomrule
\toprule
Mid Frequency Noise &	DTW	&MNet&SPGN\\\midrule
ECG5000 (w/o)&	49.4&54.8&61.4\\
ECG5000 (w/)&	46.1 (-3.3)&	51.7 (-3.1)&	60.6 (\textbf{-0.8})\\
Cricket\_Z (w/o)&	56.0&	59.8&69.1\\
Cricket\_Z (w/)&	52.8 (-3.2)&	56.1 (-3.7)&68.5 (\textbf{-0.6})\\
Prox.Phal.TW (w/o)&	44.5&43.2&52.0\\
Prox.Phal.TW (w/)	&41.6 (-2.9)&	40.4 (-2.8)&51.2 (\textbf{-0.8})\\

\bottomrule
\toprule
Low Frequency Noise &	DTW	&MNet&SPGN\\\midrule
ECG5000 (w/o)&	49.4&	54.8&	61.4\\
ECG5000 (w/)&	45.8 (-3.6)&	51.2 (-3.6)&	60.5 (\textbf{-0.9})\\
Cricket\_Z (w/o)&	56.0&	59.8&	69.1\\
Cricket\_Z (w/)&	52.4 (-3.6)&	56.0 (-3.8)&	68.3 (\textbf{-0.8})\\
Prox.Phal.TW (w/o)&	44.5&	43.2&	52.0\\
Prox.Phal.TW (w/)&	40.7 (-3.8)&	40.8 (-2.4)&	51.1 (\textbf{-0.9})\\
\bottomrule
\end{tabular}
}
\end{table}

\subsection{The Impact of SRA and SP}
Spectral Relations Acquisition and Spectral Propagation are both important modules of SPGN. To investigate the impact of both modules quantitatively, we design experiments to by removing them firstly and then adding them respectively. For baseline experiment, we simply extract temporal feature of support and query time series and adopt a weighted nearest neighbour classifier to make predictions. We add SRA module to baseline by expanding time series in power spectral space and make spectrum-wise comparisons without label propagation while we add SP module to baseline through just concatenating temporal feature with spectral feature without spectral relations extraction in graph network. All the experiments are conducted on ECG5000 and the results are shown in Table \ref{ablation}. The SRA module improves the performance over baseline by 3.7\% which illustrates the necessity of spectrum-wise relations for few-shot TSC tasks. The SP module promotes the classification accuracy by 3.2\% and shows the full exploitation of fine-grained spectral differences. It is noted that the accuracy improvement with combining SRA with SP surpasses the summation accuracy of only using SRA and SP respectively. This phenomenon further conveys that there exists a bi-directional effect between SRA and SP and they promotes each other mutually. Additionally, the number $l$ of graph networks existing in SPGN also has an impact on few-shot TSC tasks. We experiment on UCR datasets to find a proper $l$ and results reveal that SPGN achieves best performance when $l$ is set to $3$. 
\begin{table}[ht]
  \caption{Ablation study results of Spectral Relations Acquisition and Spectral Propagation modules on ECG5000 dataset. }
  \label{ablation}
  \centering
  \begin{tabular}{lll|l}
    \toprule
    baseline & +SRA & +SP & accuracy\\ 
    \midrule
    \checkmark&   &  & 53.2   \\ 
    \checkmark    & \checkmark & & 56.9 \textbf{(+3.7)}   \\
    \checkmark    &       & \checkmark &56.4 \textbf{(+3.2)} \\
    \checkmark    &  \checkmark   & \checkmark & 61.4 \textbf{(+8.2)}\\
    \bottomrule
 \end{tabular}
 \end{table}
\paragraph{The Number of Support Samples} We conduct the experiments to investigate the accuracy trend with the number K of support samples changing. Table \ref{shot} demonstrates that the average rank is better with the K increasing and our SPGN outperforms other methods in all cases. It further prove the generalization ability of our proposed SPGN in few-shot scenarios.
\begin{table}[ht]
    \centering
        \caption{The trend of average rank with the number K of support samples increasing.}
    \label{shot}
    \begin{tabular}{l|cccc}
    \toprule
K&	DTW&	BOSS&	MNet&	\textbf{SPGN}\\
\midrule
2&	3.976&	3.902&	3.178&\textbf{2.209}\\
5&	3.503&	3.890&	2.961&	\textbf{2.156}\\
10&	3.476&	3.646&	2.785&	\textbf{1.953}\\
20&	3.354&	2.927&	2.695&\textbf{1.884}\\
\bottomrule
\end{tabular}
\end{table}


\begin{table}
\centering
    \caption{Accuracies comparisons between different split methods.}
    \label{way}
    \begin{tabular}{llll}
        \toprule
        Datasets & Exponential & Average & \textbf{Ours}\\ 
        \midrule
        Adiac&  75.2 &  74.3&  \textbf{77.6}  \\ 
        Beef&81.4&81.1&\textbf{83.5}\\
        Meat&98.0&96.9&\textbf{99.1}\\
        Yoga&61.7&61.4&\textbf{63.5}\\
        Wine&71.3&70.8&\textbf{72.3}\\
        SwedishLeaf&86.6&85.9&\textbf{88.1}\\
        FaceAll &  83.1 & 81.9 &    \textbf{85.2} \\
        ECG5000   & 58.5&     57.0  & \textbf{61.4}  \\
        FordB  &  84.0&  82.7 & \textbf{87.1} \\
        50words &   69.7&    68.5 & \textbf{72.3}\\
        \bottomrule
    \end{tabular}
\end{table}

\begin{figure}
    \centering
    \includegraphics[width=6cm]{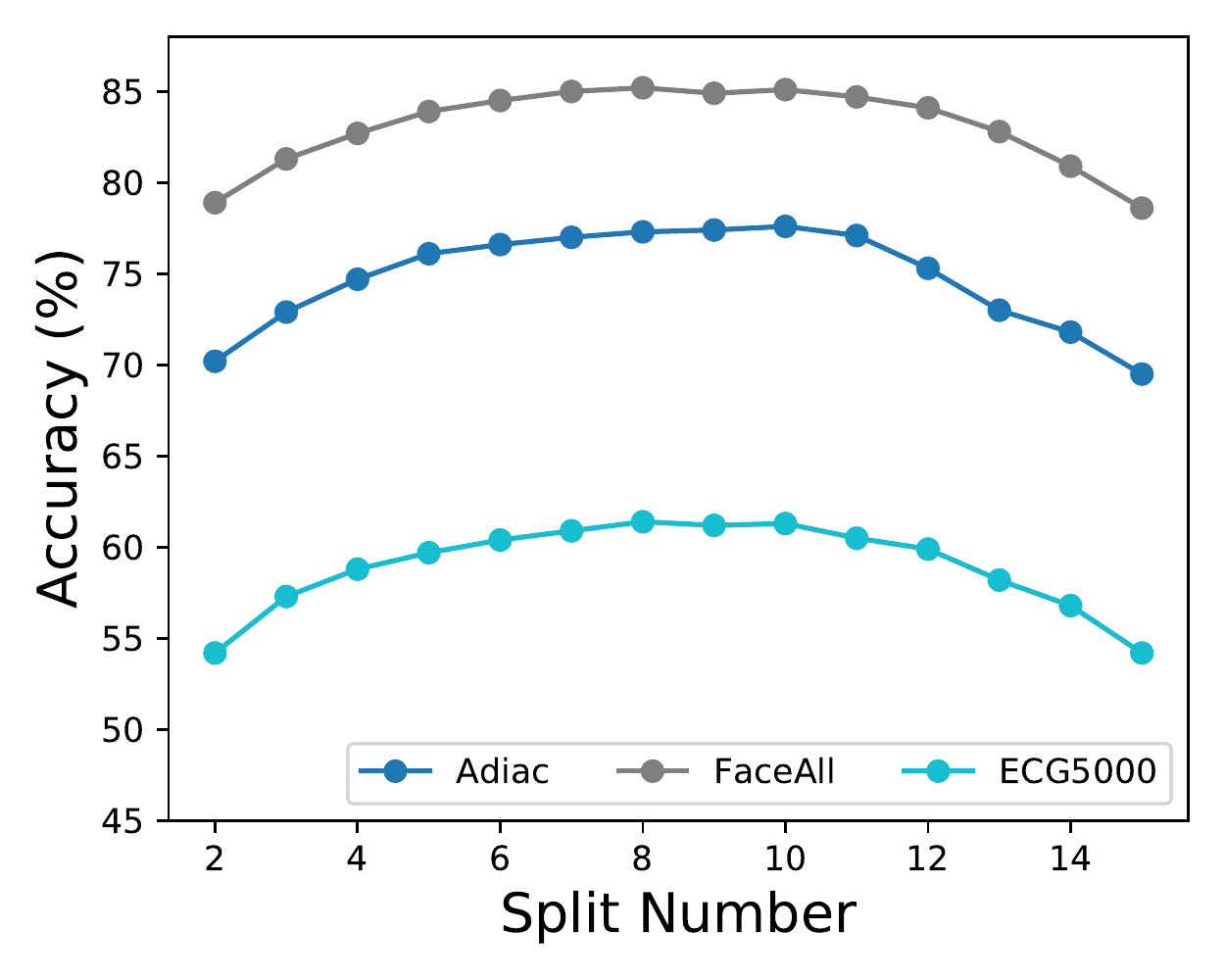}
    \caption{The trend of accuracies with the split number increasing.}
    \label{fig:split}
\end{figure}

\subsection{Investigation of Spectrum Split}
As illustrated in \ref{split}, we split spectrum of time series in an average way on power spectral space. We have conducted experiments to investigate the different method of spectral split in few-shot TSC tasks. Results depicted in Table \ref{way} shows that our method of  conducting average split on power spectral space outperforms the approaches of exponential and average split on frequency, which acquires the most accurate predictions for query time series across datasets. In addition, we explore the trend of accuracy with the number of intervals changing. Specifically, we vary the split number from 2 to 15 and evaluate the model on few-shot TSC tasks. From Figure \ref{fig:split}, we could see that SPGN achieves higher accuracy with the number increasing from 2 to 8 because of the richer spectral relations. The accuracy nearly remains unchanged from 8 to 10 and begins to decline quickly from 10 to 15. The reason for this phenomenon is that too many spectral splits are not only unhelpful but also confusing for few-shot TSC, which bring in more confounders relations such as noise, variances within same class and common feature between time series of different classes. 






\section{Conclusion and Future Work}
\label{conclusion}
In this paper, we presented Spectral Propagation Graph Network for few-shot time series classification, which is the first to leverage spectrum-wise relations and involve label propagation with spectral information in graph networks. The further study and analysis of the proposed model prove the superiority of SPGN. Extensive experiments show that SPGN achieve the state-of-the-art results on few-shot TSC tasks and outperforms previous methods by a significant margin from different aspects, such as cross-domain and cross-way evaluation. However, the split number of SPGN is acquired with many trials across datasets, which is complicated for practice.
In the future work, we aim to use Maximum Likelihood Estimation (MLE) to obtain a proper split number in advance and extend our model to other time series tasks.


\bibliography{nips}

\begin{thebibliography}{54}
\providecommand{\natexlab}[1]{#1}
\providecommand{\url}[1]{\texttt{#1}}
\expandafter\ifx\csname urlstyle\endcsname\relax
  \providecommand{\doi}[1]{doi: #1}\else
  \providecommand{\doi}{doi: \begingroup \urlstyle{rm}\Url}\fi

\bibitem[Agarap(2018)]{agarap2018deep}
Abien~Fred Agarap.
\newblock Deep learning using rectified linear units (relu).
\newblock \emph{arXiv preprint arXiv:1803.08375}, 2018.

\bibitem[Bagnall et~al.(2017)Bagnall, Lines, Bostrom, Large, and
  Keogh]{bagnall2017great}
Anthony Bagnall, Jason Lines, Aaron Bostrom, James Large, and Eamonn Keogh.
\newblock The great time series classification bake off: a review and
  experimental evaluation of recent algorithmic advances.
\newblock \emph{Data mining and knowledge discovery}, 31\penalty0 (3):\penalty0
  606--660, 2017.

\bibitem[Chen et~al.(2018)Chen, Liu, Kira, Wang, and Huang]{chen2018closer}
Wei-Yu Chen, Yen-Cheng Liu, Zsolt Kira, Yu-Chiang~Frank Wang, and Jia-Bin
  Huang.
\newblock A closer look at few-shot classification.
\newblock In \emph{International Conference on Learning Representations}, 2018.

\bibitem[Dau et~al.(2019)Dau, Bagnall, Kamgar, Yeh, Zhu, Gharghabi,
  Ratanamahatana, and Keogh]{dau2019ucr}
Hoang~Anh Dau, Anthony Bagnall, Kaveh Kamgar, Chin-Chia~Michael Yeh, Yan Zhu,
  Shaghayegh Gharghabi, Chotirat~Ann Ratanamahatana, and Eamonn Keogh.
\newblock The ucr time series archive.
\newblock \emph{IEEE/CAA Journal of Automatica Sinica}, 6\penalty0
  (6):\penalty0 1293--1305, 2019.

\bibitem[Esling \& Agon(2012)Esling and Agon]{esling2012time}
Philippe Esling and Carlos Agon.
\newblock Time-series data mining.
\newblock \emph{ACM Computing Surveys (CSUR)}, 45\penalty0 (1):\penalty0 1--34,
  2012.

\bibitem[Fawaz et~al.(2018)Fawaz, Forestier, Weber, Idoumghar, and
  Muller]{fawaz2018transfer}
Hassan~Ismail Fawaz, Germain Forestier, Jonathan Weber, Lhassane Idoumghar, and
  Pierre-Alain Muller.
\newblock Transfer learning for time series classification.
\newblock In \emph{2018 IEEE international conference on big data (Big Data)},
  pp.\  1367--1376. IEEE, 2018.

\bibitem[Fawaz et~al.(2019)Fawaz, Forestier, Weber, Idoumghar, and
  Muller]{fawaz2019deep}
Hassan~Ismail Fawaz, Germain Forestier, Jonathan Weber, Lhassane Idoumghar, and
  Pierre-Alain Muller.
\newblock Deep learning for time series classification: a review.
\newblock \emph{Data Mining and Knowledge Discovery}, 33\penalty0 (4):\penalty0
  917--963, 2019.

\bibitem[Finn et~al.(2017)Finn, Abbeel, and Levine]{finn2017model}
Chelsea Finn, Pieter Abbeel, and Sergey Levine.
\newblock Model-agnostic meta-learning for fast adaptation of deep networks.
\newblock In \emph{International Conference on Machine Learning}, pp.\
  1126--1135. PMLR, 2017.

\bibitem[Gori et~al.(2005)Gori, Monfardini, and Scarselli]{gori2005new}
Marco Gori, Gabriele Monfardini, and Franco Scarselli.
\newblock A new model for learning in graph domains.
\newblock In \emph{Proceedings. 2005 IEEE International Joint Conference on
  Neural Networks, 2005.}, volume~2, pp.\  729--734. IEEE, 2005.

\bibitem[Hatami et~al.(2018)Hatami, Gavet, and
  Debayle]{hatami2018classification}
Nima Hatami, Yann Gavet, and Johan Debayle.
\newblock Classification of time-series images using deep convolutional neural
  networks.
\newblock In \emph{Tenth international conference on machine vision (ICMV
  2017)}, volume 10696, pp.\  106960Y. International Society for Optics and
  Photonics, 2018.

\bibitem[He et~al.(2016)He, Zhang, Ren, and Sun]{he2016deep}
Kaiming He, Xiangyu Zhang, Shaoqing Ren, and Jian Sun.
\newblock Deep residual learning for image recognition.
\newblock In \emph{Proceedings of the IEEE conference on computer vision and
  pattern recognition}, pp.\  770--778, 2016.

\bibitem[Hong et~al.(2020)Hong, Xu, Khare, Priambada, Maher, Aljiffry, Sun, and
  Tumanov]{hong2020holmes}
Shenda Hong, Yanbo Xu, Alind Khare, Satria Priambada, Kevin Maher, Alaa
  Aljiffry, Jimeng Sun, and Alexey Tumanov.
\newblock Holmes: Health online model ensemble serving for deep learning models
  in intensive care units.
\newblock In \emph{Proceedings of the 26th ACM SIGKDD International Conference
  on Knowledge Discovery \& Data Mining}, pp.\  1614--1624, 2020.

\bibitem[Huikuri et~al.(2001)Huikuri, Castellanos, and
  Myerburg]{huikuri2001sudden}
Heikki~V Huikuri, Agustin Castellanos, and Robert~J Myerburg.
\newblock Sudden death due to cardiac arrhythmias.
\newblock \emph{New England Journal of Medicine}, 345\penalty0 (20):\penalty0
  1473--1482, 2001.

\bibitem[Ioffe \& Szegedy(2015)Ioffe and Szegedy]{ioffe2015batch}
Sergey Ioffe and Christian Szegedy.
\newblock Batch normalization: Accelerating deep network training by reducing
  internal covariate shift.
\newblock In \emph{International conference on machine learning}, pp.\
  448--456. PMLR, 2015.

\bibitem[Joachims et~al.(1999)]{joachims1999transductive}
Thorsten Joachims et~al.
\newblock Transductive inference for text classification using support vector
  machines.
\newblock In \emph{Icml}, volume~99, pp.\  200--209, 1999.

\bibitem[Kashiparekh et~al.(2019)Kashiparekh, Narwariya, Malhotra, Vig, and
  Shroff]{kashiparekh2019convtimenet}
Kathan Kashiparekh, Jyoti Narwariya, Pankaj Malhotra, Lovekesh Vig, and Gautam
  Shroff.
\newblock Convtimenet: A pre-trained deep convolutional neural network for time
  series classification.
\newblock In \emph{2019 International Joint Conference on Neural Networks
  (IJCNN)}, pp.\  1--8. IEEE, 2019.

\bibitem[Kim et~al.(2019)Kim, Kim, Kim, and Yoo]{kim2019edge}
Jongmin Kim, Taesup Kim, Sungwoong Kim, and Chang~D Yoo.
\newblock Edge-labeling graph neural network for few-shot learning.
\newblock In \emph{Proceedings of the IEEE/CVF Conference on Computer Vision
  and Pattern Recognition}, pp.\  11--20, 2019.

\bibitem[Kingma \& Ba(2015)Kingma and Ba]{kingma2015adam}
Diederik~P Kingma and Jimmy Ba.
\newblock Adam: A method for stochastic optimization.
\newblock In \emph{ICLR (Poster)}, 2015.

\bibitem[Krizhevsky et~al.(2012)Krizhevsky, Sutskever, and
  Hinton]{krizhevsky2012imagenet}
Alex Krizhevsky, Ilya Sutskever, and Geoffrey~E Hinton.
\newblock Imagenet classification with deep convolutional neural networks.
\newblock \emph{Advances in neural information processing systems},
  25:\penalty0 1097--1105, 2012.

\bibitem[LeCun et~al.(2015)LeCun, Bengio, and Hinton]{lecun2015deep}
Yann LeCun, Yoshua Bengio, and Geoffrey Hinton.
\newblock Deep learning.
\newblock \emph{nature}, 521\penalty0 (7553):\penalty0 436--444, 2015.

\bibitem[Lines \& Bagnall(2015)Lines and Bagnall]{lines2015time}
Jason Lines and Anthony Bagnall.
\newblock Time series classification with ensembles of elastic distance
  measures.
\newblock \emph{Data Mining and Knowledge Discovery}, 29\penalty0 (3):\penalty0
  565--592, 2015.

\bibitem[Liu et~al.(2018)Liu, Lee, Park, Kim, Yang, Hwang, and
  Yang]{liu2018learning}
Yanbin Liu, Juho Lee, Minseop Park, Saehoon Kim, Eunho Yang, Sung~Ju Hwang, and
  Yi~Yang.
\newblock Learning to propagate labels: Transductive propagation network for
  few-shot learning.
\newblock In \emph{International Conference on Learning Representations}, 2018.

\bibitem[Malhotra et~al.(2017)Malhotra, TV, Vig, Agarwal, and
  Shroff]{DBLP:journals/corr/MalhotraTVAS17}
Pankaj Malhotra, Vishnu TV, Lovekesh Vig, Puneet Agarwal, and Gautam Shroff.
\newblock Timenet: Pre-trained deep recurrent neural network for time series
  classification.
\newblock \emph{CoRR}, abs/1706.08838, 2017.
\newblock URL \url{http://arxiv.org/abs/1706.08838}.

\bibitem[Nakamura \& Harada(2019)Nakamura and Harada]{nakamura2019revisiting}
Akihiro Nakamura and Tatsuya Harada.
\newblock Revisiting fine-tuning for few-shot learning.
\newblock \emph{arXiv preprint arXiv:1910.00216}, 2019.

\bibitem[Narwariya et~al.(2020)Narwariya, Malhotra, Vig, Shroff, and
  Vishnu]{narwariya2020meta}
Jyoti Narwariya, Pankaj Malhotra, Lovekesh Vig, Gautam Shroff, and TV~Vishnu.
\newblock Meta-learning for few-shot time series classification.
\newblock In \emph{Proceedings of the 7th ACM IKDD CoDS and 25th COMAD}, pp.\
  28--36. 2020.

\bibitem[Nweke et~al.(2018)Nweke, Teh, Al-Garadi, and Alo]{nweke2018deep}
Henry~Friday Nweke, Ying~Wah Teh, Mohammed~Ali Al-Garadi, and Uzoma~Rita Alo.
\newblock Deep learning algorithms for human activity recognition using mobile
  and wearable sensor networks: State of the art and research challenges.
\newblock \emph{Expert Systems with Applications}, 105:\penalty0 233--261,
  2018.

\bibitem[Pyakillya et~al.(2017)Pyakillya, Kazachenko, and
  Mikhailovsky]{pyakillya2017deep}
B~Pyakillya, N~Kazachenko, and N~Mikhailovsky.
\newblock Deep learning for ecg classification.
\newblock In \emph{Journal of physics: conference series}, volume 913, pp.\
  012004. IOP Publishing, 2017.

\bibitem[Rajkomar et~al.(2018)Rajkomar, Oren, Chen, Dai, Hajaj, Hardt, Liu,
  Liu, Marcus, Sun, et~al.]{rajkomar2018scalable}
Alvin Rajkomar, Eyal Oren, Kai Chen, Andrew~M Dai, Nissan Hajaj, Michaela
  Hardt, Peter~J Liu, Xiaobing Liu, Jake Marcus, Mimi Sun, et~al.
\newblock Scalable and accurate deep learning with electronic health records.
\newblock \emph{NPJ Digital Medicine}, 1\penalty0 (1):\penalty0 1--10, 2018.

\bibitem[Saritha et~al.(2008)Saritha, Sukanya, and Murthy]{saritha2008ecg}
C~Saritha, V~Sukanya, and Y~Narasimha Murthy.
\newblock Ecg signal analysis using wavelet transforms.
\newblock \emph{Bulg. J. Phys}, 35\penalty0 (1):\penalty0 68--77, 2008.

\bibitem[Satorras \& Estrach(2018)Satorras and Estrach]{satorras2018few}
Victor~Garcia Satorras and Joan~Bruna Estrach.
\newblock Few-shot learning with graph neural networks.
\newblock In \emph{International Conference on Learning Representations}, 2018.

\bibitem[Scarselli et~al.(2008)Scarselli, Gori, Tsoi, Hagenbuchner, and
  Monfardini]{scarselli2008graph}
Franco Scarselli, Marco Gori, Ah~Chung Tsoi, Markus Hagenbuchner, and Gabriele
  Monfardini.
\newblock The graph neural network model.
\newblock \emph{IEEE transactions on neural networks}, 20\penalty0
  (1):\penalty0 61--80, 2008.

\bibitem[Sch{\"a}fer(2015)]{schafer2015boss}
Patrick Sch{\"a}fer.
\newblock The boss is concerned with time series classification in the presence
  of noise.
\newblock \emph{Data Mining and Knowledge Discovery}, 29\penalty0 (6):\penalty0
  1505--1530, 2015.

\bibitem[Schroff et~al.(2015)Schroff, Kalenichenko, and
  Philbin]{schroff2015facenet}
Florian Schroff, Dmitry Kalenichenko, and James Philbin.
\newblock Facenet: A unified embedding for face recognition and clustering.
\newblock In \emph{Proceedings of the IEEE conference on computer vision and
  pattern recognition}, pp.\  815--823, 2015.

\bibitem[Serr{\`a} et~al.(2018)Serr{\`a}, Pascual, and
  Karatzoglou]{serra2018towards}
Joan Serr{\`a}, Santiago Pascual, and Alexandros Karatzoglou.
\newblock Towards a universal neural network encoder for time series.
\newblock In \emph{CCIA}, pp.\  120--129, 2018.

\bibitem[Snell et~al.(2017)Snell, Swersky, and Zemel]{snell2017prototypical}
Jake Snell, Kevin Swersky, and Richard Zemel.
\newblock Prototypical networks for few-shot learning.
\newblock In \emph{Proceedings of the 31st International Conference on Neural
  Information Processing Systems}, pp.\  4080--4090, 2017.

\bibitem[Sundermeyer et~al.(2012)Sundermeyer, Schl{\"u}ter, and
  Ney]{sundermeyer2012lstm}
Martin Sundermeyer, Ralf Schl{\"u}ter, and Hermann Ney.
\newblock Lstm neural networks for language modeling.
\newblock In \emph{Thirteenth annual conference of the international speech
  communication association}, 2012.

\bibitem[Sung et~al.(2018)Sung, Yang, Zhang, Xiang, Torr, and
  Hospedales]{sung2018learning}
Flood Sung, Yongxin Yang, Li~Zhang, Tao Xiang, Philip~HS Torr, and Timothy~M
  Hospedales.
\newblock Learning to compare: Relation network for few-shot learning.
\newblock In \emph{Proceedings of the IEEE conference on computer vision and
  pattern recognition}, pp.\  1199--1208, 2018.

\bibitem[Vilalta \& Drissi(2002)Vilalta and Drissi]{vilalta2002perspective}
Ricardo Vilalta and Youssef Drissi.
\newblock A perspective view and survey of meta-learning.
\newblock \emph{Artificial intelligence review}, 18\penalty0 (2):\penalty0
  77--95, 2002.

\bibitem[Vinyals et~al.(2016)Vinyals, Blundell, Lillicrap, Kavukcuoglu, and
  Wierstra]{vinyals2016matching}
Oriol Vinyals, Charles Blundell, Timothy Lillicrap, Koray Kavukcuoglu, and Daan
  Wierstra.
\newblock Matching networks for one shot learning.
\newblock \emph{arXiv preprint arXiv:1606.04080}, 2016.

\bibitem[Wang \& Zhang(2007)Wang and Zhang]{wang2007label}
Fei Wang and Changshui Zhang.
\newblock Label propagation through linear neighborhoods.
\newblock \emph{IEEE Transactions on Knowledge and Data Engineering},
  20\penalty0 (1):\penalty0 55--67, 2007.

\bibitem[Wang et~al.(2020)Wang, Yao, Kwok, and Ni]{wang2020generalizing}
Yaqing Wang, Quanming Yao, James~T Kwok, and Lionel~M Ni.
\newblock Generalizing from a few examples: A survey on few-shot learning.
\newblock \emph{ACM Computing Surveys (CSUR)}, 53\penalty0 (3):\penalty0 1--34,
  2020.

\bibitem[Wang \& Oates(2015{\natexlab{a}})Wang and Oates]{wang2015imaging}
Zhiguang Wang and Tim Oates.
\newblock Imaging time-series to improve classification and imputation.
\newblock In \emph{IJCAI}, 2015{\natexlab{a}}.

\bibitem[Wang \& Oates(2015{\natexlab{b}})Wang and Oates]{wang2015spatially}
Zhiguang Wang and Tim Oates.
\newblock Spatially encoding temporal correlations to classify temporal data
  using convolutional neural networks.
\newblock \emph{arXiv preprint arXiv:1509.07481}, 2015{\natexlab{b}}.

\bibitem[Wang et~al.(2017)Wang, Yan, and Oates]{wang2017time}
Zhiguang Wang, Weizhong Yan, and Tim Oates.
\newblock Time series classification from scratch with deep neural networks: A
  strong baseline.
\newblock In \emph{2017 International joint conference on neural networks
  (IJCNN)}, pp.\  1578--1585. IEEE, 2017.

\bibitem[Wellens et~al.(2000)Wellens, Schuilenburg, and
  Durrer]{wellens2000electrical}
Hein~JJ Wellens, Reinier~M Schuilenburg, and Dirk Durrer.
\newblock Electrical stimulation of the heart in patients with ventricular
  tachycardia.
\newblock In \emph{Professor Hein JJ Wellens}, pp.\  43--53. Springer, 2000.

\bibitem[Xu et~al.(2022{\natexlab{a}})Xu, Tu, Xiang, Shao, Zhou, and
  Ma]{xu2022cobevt}
Runsheng Xu, Zhengzhong Tu, Hao Xiang, Wei Shao, Bolei Zhou, and Jiaqi Ma.
\newblock Cobevt: Cooperative bird's eye view semantic segmentation with sparse
  transformers.
\newblock \emph{arXiv preprint arXiv:2207.02202}, 2022{\natexlab{a}}.

\bibitem[Xu et~al.(2022{\natexlab{b}})Xu, Xiang, Tu, Xia, Yang, and
  Ma]{xu2022v2x}
Runsheng Xu, Hao Xiang, Zhengzhong Tu, Xin Xia, Ming-Hsuan Yang, and Jiaqi Ma.
\newblock V2x-vit: Vehicle-to-everything cooperative perception with vision
  transformer.
\newblock \emph{arXiv preprint arXiv:2203.10638}, 2022{\natexlab{b}}.

\bibitem[Yang \& Hong(2022{\natexlab{a}})Yang and Hong]{pmlr-v162-yang22d}
Ling Yang and Shenda Hong.
\newblock Omni-granular ego-semantic propagation for self-supervised graph
  representation learning.
\newblock In \emph{Proceedings of the 39th International Conference on Machine
  Learning}, pp.\  25022--25037. PMLR, 2022{\natexlab{a}}.

\bibitem[Yang \& Hong(2022{\natexlab{b}})Yang and Hong]{yang2022omni}
Ling Yang and Shenda Hong.
\newblock Omni-granular ego-semantic propagation for self-supervised graph
  representation learning.
\newblock \emph{arXiv preprint arXiv:2205.15746}, 2022{\natexlab{b}}.

\bibitem[Yang \& Hong(2022{\natexlab{c}})Yang and Hong]{yang2022unsupervised}
Ling Yang and Shenda Hong.
\newblock Unsupervised time-series representation learning with iterative
  bilinear temporal-spectral fusion.
\newblock In \emph{Proceedings of the 39th International Conference on Machine
  Learning}, pp.\  25038--25054. PMLR, 2022{\natexlab{c}}.

\bibitem[Yang \& Hong(2022{\natexlab{d}})Yang and
  Hong]{yang2022unsupervised-arxiv}
Ling Yang and Shenda Hong.
\newblock Unsupervised time-series representation learning with iterative
  bilinear temporal-spectral fusion.
\newblock \emph{arXiv preprint arXiv:2202.04770}, 2022{\natexlab{d}}.

\bibitem[Yang et~al.(2020)Yang, Li, Zhang, Zhou, Zhou, and Liu]{yang2020dpgn}
Ling Yang, Liangliang Li, Zilun Zhang, Xinyu Zhou, Erjin Zhou, and Yu~Liu.
\newblock Dpgn: Distribution propagation graph network for few-shot learning.
\newblock In \emph{Proceedings of the IEEE/CVF Conference on Computer Vision
  and Pattern Recognition}, pp.\  13390--13399, 2020.

\bibitem[Yang \& Wu(2006)Yang and Wu]{yang200610}
Qiang Yang and Xindong Wu.
\newblock 10 challenging problems in data mining research.
\newblock \emph{International Journal of Information Technology \& Decision
  Making}, 5\penalty0 (04):\penalty0 597--604, 2006.

\bibitem[Zhou et~al.(2004)Zhou, Bousquet, Lal, Weston, and
  Sch{\"o}lkopf]{zhou2004learning}
Dengyong Zhou, Olivier Bousquet, Thomas~N Lal, Jason Weston, and Bernhard
  Sch{\"o}lkopf.
\newblock Learning with local and global consistency.
\newblock In \emph{Advances in neural information processing systems}, pp.\
  321--328, 2004.

\end{thebibliography}
\bibliographystyle{ieee}

\end{document}